\documentclass{article}
\usepackage{ijcai22}

\usepackage{times}
\usepackage{soul}
\usepackage{url}
\usepackage[hidelinks]{hyperref}
\usepackage[utf8]{inputenc}
\usepackage[small]{caption}
\usepackage{graphicx}
\usepackage{amsmath}
\usepackage{amsthm}
\usepackage{booktabs}
\usepackage{algorithm}
\usepackage{algorithmic}
\newcommand{\temprel}[1]{$\langle$\textit{#1}$\rangle$}

\title{Controllable Text Generation for Open-Domain Creativity and Fairness}

\author{
    Nanyun Peng
    \affiliations
    Computer Science Department, 
  University of California, Los Angeles
    \emails
    violetpeng@cs.ucla.edu
}

\begin{document}

\maketitle

\begin{abstract}
    Recent advances in large pre-trained language models have demonstrated strong results in generating natural languages and significantly improved performances for many natural language generation (NLG) applications such as machine translation and text summarization. However, when the generation tasks are more open-ended and the content is under-specified, existing techniques struggle to generate long-term coherent and creative content. Moreover, the models exhibit and even amplify social biases that are learned from the training corpora. This happens because the generation models are trained to capture the surface patterns (i.e. sequences of words), instead of capturing underlying semantics and discourse structures, as well as background knowledge including social norms. In this paper, I introduce our recent works on controllable text generation to enhance the creativity and fairness of language generation models. We explore hierarchical generation and constrained decoding, with applications to creative language generation including story, poetry, and figurative languages, and bias mitigation for generation models.
\end{abstract}

\section{Introduction}
Natural language generation (NLG) tasks can be mapped into a spectrum based on their conditional entropy, i.e., the uncertainty of the output distribution given the inputs. One side of the spectrum is the low conditional entropy tasks, such as machine translation, abstractive summarization, and task-oriented dialogue systems, where the inputs largely determine the contents of the outputs. 
On the other side is the tasks that are open-ended and have high conditional entropy, such as story, poetry, or lyric generation given a title or a prompt, and open-domain dialogue systems (chit-chat), dubbed \textit{creative generation}. In addition to be able to compose grammatical and fluence sentences to articulate given contents, these tasks usually also require extensive world and common sense knowledge, discourse-level coherence modeling to make sure the outputs are long-term coherent, sensible, and creative, making them especially challenging NLG tasks. 

To tackle the challenges, my group has been working on several synergistic directions to push the frontier of open-domain (creative) generation, with a shared modeling theme of controllable text generation and applications to story, poetry, and dialogue response generation. 

The first direction is to build hierarchical models that disentangle the content planning (plan out the structured representation of the content) from surface realization (convert the structured content to natural language sentences). This has three major benefits: i). It enhances the controllability of the generation through the plan and humans can more easily collaborate with the models on the plan-level to create novel contents. ii). It improves long-term conherence of the generation, since the compact plan representation helps the model to capture the high-level outline of the generation target, and iii). It enables the introduction of creativity on both the planning-level (e.g., better events and their temporal and causal relation modeling to introduce surprise) and the diction-level (e.g., using figurative language to improve the writing). More details about this direction will be discussed in Section~\ref{sec:hierarchical}.
Additionally, we develop fundamental models to support the controllable, hierarchical generation pipeline. i) We design insertion-based models that break the left-to-right generation order to better incorporate constraints/control factors (more details in Section~\ref{sec:controllable}). ii)
We propose various constrained decoding algorithms to control the models.
Finally, we have also been working on analyzing and reducing biases in open-domain generation (more details in Section~\ref{sec:bias}).


\section{Hierarchical Text Generation}
\label{sec:hierarchical}
Hierarchical text generation follows the \textit{plan-and-write}~\cite{yao2019plan} paradigm to first plan out the main content of the output, and then convert these structured plans into natural language texts (also called surface realization).
The idea of content planning for text generation dates back to the late 70s~\cite{meehan1976metanovel} with various plan representations. 
Prior works predominately rely on \textit{manually} constructed or \textit{rule-based} plans, which are either expensive to construct, or restricted to specific domains. 

\subsection{Automatic Content Plan Extraction}
My group has been exploring the direction of learning content planning from existing corpora by using information extraction tools to extract ``silver standard'' plan representations from data for models to learn planning~\cite{yao2019plan,goldfarb2020content,han2022go}. 
Specifically, we have extensive works on extracting entities~\cite{huang2019learning,huang2021tempgen}, entity relations~\cite{Peng16graphlstm,hsu2021discourse}, events~\cite{ahmad2021gate,hsu2022degree}, and event temporal relations~\cite{han2019joint,han2020knowledge,han2021econet}
from text corpora, to compose ``silver standard'' content plans for models to learn and generate novel plans during the test time. 

\subsection{Content Planning with Temporal Modeling, Literary Principles, and Surprise}
While content plans have shown to be effective in improving the coherence of the generation~\cite{yao2019plan}, prior works also have shown that generated plans are much lower quality than the silver plans as they are usually repetitive, boring, and violate common sense and world knowledge~\cite{martin2017event,yao2019plan,Fan2019StrategiesFS}. 
To this end, we have been working on  improving the planning model by introducing knowledge about literary principles~\cite{goldfarb2020content}, event temporal knowledge~\cite{han2022go}, and common sense knowledge. 
Specifically, we represent the plan generation model as a probabilistic model $p(\textbf{z}|\textbf{x})$ that is fitted by auto-regressive generative neural networks, where $\bf{z}$ represents the structured plan, $\bf{x}$ represents a given prompt. 
We modify the decoding objective to incorporate several rescoring models $a$ $\in$ $A$ to re-rank the original, or ``naive" plans  generated by the graph generative model. These rescoring models bring the generated plan graph closer to each rescoring model's specialty (such as relevance, coherence, and temporality). The modified decoding objective becomes:
$\begin{aligned} 
f_\lambda({\bf x, z}) = \sum_{i}^{m}-\log p(z|z<i,{\bf x}) + \sum_{j}^{|A|} \lambda_{j}a_j({\bf x}, z_{i...m}) 
\end{aligned}$
where $\lambda_j$ is the learned weight of the score given by $a_j$. What differs for each model $a_j$ that specializes in a different rescoring aspect is the set of training data we generated to train the discriminators. 

We also explored improving the interestingness of the generation, by introducing temporal diversity to the events, such as \textit{flashbacks} that insert past events into current storylines as we commonly observe in novels and plays~\cite{han2022go}. It is challenging for machines to generate \textit{flashbacks} as it requires solid understanding of event \textbf{temporal order} (e.g. \textit{feeling hungry} \temprel{before} \textit{eat}, not vice versa), and the creativity to arrange storylines so that earlier events do not always appear first in \textbf{narrative order}. 
Two major issues in existing systems exacerbate the challenges: 1) temporal bias in pretraining and datasets that leads to monotonic event temporal orders; 2) lack of explicit guidance that helps machines decide where to insert \textit{flashbacks}. We address these issues by introducing \textit{temporal prompts} to the structured plans to encode events and their pair-wise temporal relations (\temprel{before}, \temprel{after} and \temprel{vague}) to guide how stories should unfold temporally. We showed that temporal prompts helped generate more interesting stories with \textit{flashbacks} while maintaining textual diversity, fluency and temporal coherence.


\subsection{Introducing Literary Aesthetics in Writing}
Given a well-crafted plan, the other important component for the hierarchical generation is surface realization that converts plans into natural language sentences. The state-of-the-art pre-trained language models~\cite{radford2019language,lewis2020bart} are capable of generating grammatical and fluent sentences, but they are usually plain and dull. To improve aesthetics in writing, we have explored incorporating figurative languages into the generation. 

Figures of speech are literary devices that are commonly employed in narratives and stories~\cite{boudens2005story}. Proper usage of figurative languages can improve the effectiveness, interestingness, and enjoyment of communications~\cite{stein2018using}. 
However, the composition of figures of speech usually requires extensive contextual, common sense, and world knowledge~\cite{justo2014extracting,yoshimura2015simile}, which remains an open challenge for NLG. My group has been working on generating puns, similes, sarcasms, and metaphors ~\cite{he2019pun,chakrabarty2020r,chakrabarty2020generating,chakrabarty2021mermaid,stowe2021metaphor,Mittal2022ambipun}, and incorporate them into story~\cite{chakrabarty2020generating} and poetry~\cite{chakrabarty2021mermaid,tian2022sonnet} generation to successfully improve the aesthetics of the writing. 

\section{Controllable Text Generation}
\label{sec:controllable}
One advantage of the hierarchical generation framework is the improved controllability of the generation through the plan -- human writers can more easily collaborate with the models on the plan-level to control the generation contents. However, it is not straightforward to incorporate such controls in the state-of-the-art language generation models.
The pre-trained auto-regressive language models~\cite{radford2019language,lewis2020bart} usually generate sentences word by word from left to right, making the control of the generation process challenging. 

\subsection{Insertion-based Generation}
Prior works~\cite{Fan2019StrategiesFS,yao2019plan} observed that fine-tune sequence-to-sequence models, while works decent for controllable language generate, cannot guarantee faithful incorporation of the plan. 
For example, we~\cite{yao2019plan} showed that there are slightly less than $80\%$ of storyline keywords (a form of plan) appeared in the generated story. 
~\cite{Fan2019StrategiesFS} thus designed special verb-attention mechanism and incorporated copy mechanism to enhance the incorporation of the plans in the generation outputs.
In light of these observations, we have been exploring a fundamentally different direction of controllable text generation framework -- insertion-based text generation, or text infilling. 


We proposed a general algorithm for \textit{efficient} insertion-based text generation~\cite{lu2021efficient} to train a permutation language model with a delicate design of the permutations to reflect the insertion orders. To suit the non-monotonic nature of the insertion-based generation process, a modified relative positional encoding mechanism is introduced such that each token is only aware of its relative position with respect to the generated partial sequence, but not the relative position with respect to the complete sequence. We showed the training efficiency, controllability, and the advantages of the model to generalize over partial observations. 
We plan to explore other applications for this novel \textit{efficient} insertion-based NLG formulation and build a large-scale pre-trained insertion-based NLG model for lexically constrained text generation.

\subsection{Controllable Decoding}
There are also methods to control auto-regressive language models to leverage the powerful pretraining. They can be summarized into three major categories: \emph{fine-tuning}, \emph{refact/retraining}, and \emph{post-processing}. The first two categories are usually inefficient considering the size of language models (billions of parameters). 
Therefore, my group has been mostly developingg post-processing based models to control pre-trained auto-regressive language models.
We have been pushing two lines of research: 1) adapting the decoding algorithm~\cite{sheng2021nice} and 2) using auxiliary models to guide the decoding~\cite{goldfarb2020content,chakrabarty2021mermaid}. 

\section{Evaluating and Mitigating Biases in Open-Domain Text Generation}
\label{sec:bias}
Evaluating open-domain generation is a challenging task as there are numerous plausible outputs given an input. Bias and fairness issues which are often subtle, should also be taken into consideration to guardrail the open-domain creative generation. 
We pioneered and have been actively exploring automatic evaluation of fairness aspects of the open-domain generation, including social stereotypes~\cite{sheng2019woman,sheng2020towards} and ad-hominem (a type of micro-aggression that is considered as toxic language)~\cite{sheng2021nice} of open-domain generation. Some of our controllable text generation models are also developed under this context to reduce biases in generated texts~\cite{sheng2020towards,sheng2021nice}.


\section{Conclusion and Future Work}
There are many exciting ongoing research directions in my group to push the frontier of natural language understanding and generation, with focused applications to creative and controllable text generation. 
Due to the space limit, I only briefly highlight three of them: 1) We are developing generation-based model for information extraction tasks. This can be applied to general natural language understanding, as well as better plan extraction for hierarchical generation. 2) We are trying to incorporate common sense knowledge into the control using common sense knowledge bases, semantic loss, and lexically constrained generation. and 3) We are diving deeper into insertion-based text generation and pre-train the first insertion-based language model on large corpora. 

\section*{Acknowledgments}
My research has been generously supported by DARPA CwC, MCS, and KAIROS programs, the IARPA BETTER program, an NIH R01 grant, and several industrial partners including CISCO, Amazon Alexa AI, and JP Morgan Chase.
I am indebted for my advisors, mentors, senior collaborators for their help, support, and coach. 
The works presented in this paper cannot be done without my PhD students and other coauthors.
Finally, I thank IJCAI 2021 program committee for the invitation to
speak at the early-career spotlight.

\bibliographystyle{named}
\bibliography{ijcai22}

\begin{thebibliography}{}

\bibitem[\protect\citeauthoryear{Ahmad \bgroup \em et al.\egroup
  }{2021}]{ahmad2021gate}
Wasi Ahmad, Nanyun Peng, and Kai-Wei Chang.
\newblock Gate: Graph attention transformer encoder for cross-lingual relation
  and event extraction.
\newblock In {\em The Thirty-Fifth AAAI Conference on Artificial Intelligence
  (AAAI-21)}, 2021.

\bibitem[\protect\citeauthoryear{Bender \bgroup \em et al.\egroup
  }{2021}]{bender2021dangers}
E.~M. Bender, T.~Gebru, A.~McMillan-Major, and S.~Shmitchell.
\newblock On the dangers of stochastic parrots: Can language models be too big?
\newblock In {\em ACM Conference on Fairness, Accountability, and Transparency
  (FAccT)}, 2021.

\bibitem[\protect\citeauthoryear{Blodgett \bgroup \em et al.\egroup
  }{2020}]{blodgett2020language}
S.~L. Blodgett, S.~Barocas, H.~{Daume III}, and H.~Wallach.
\newblock Language (technology) is power: A critical survey of "bias" in {NLP}.
\newblock In {\em Association for Computational Linguistics (ACL)}, 2020.

\bibitem[\protect\citeauthoryear{Boudens}{2005}]{boudens2005story}
Connie~J Boudens.
\newblock The story of work: A narrative analysis of workplace emotion.
\newblock {\em Organization Studies}, 26(9):1285--1306, 2005.

\bibitem[\protect\citeauthoryear{Chakrabarty \bgroup \em et al.\egroup
  }{2020a}]{chakrabarty2020r}
Tuhin Chakrabarty, Debanjan Ghosh, Smaranda Muresan, and Nanyun Peng.
\newblock $r^3$: Reverse, retrieve, and rank for sarcasm generation with
  commonsense knowledge.
\newblock In {\em ACL}, 2020.

\bibitem[\protect\citeauthoryear{Chakrabarty \bgroup \em et al.\egroup
  }{2020b}]{chakrabarty2020generating}
Tuhin Chakrabarty, Smaranda Muresan, and Nanyun Peng.
\newblock Generating similes effortlessly like a pro: A style transfer approach
  for simile generation.
\newblock In {\em Proceedings of the 2020 Conference on Empirical Methods in
  Natural Language Processing (EMNLP)}, 2020.

\bibitem[\protect\citeauthoryear{Chakrabarty \bgroup \em et al.\egroup
  }{2021}]{chakrabarty2021mermaid}
Tuhin Chakrabarty, Xurui Zhang, Smaranda Muresan, and Nanyun Peng.
\newblock Mermaid: Metaphor generation with symbolism and discriminative
  decoding.
\newblock In {\em The 2021 Annual Conference of the North American Chapter of
  the Association for Computational Linguistics (NAACL)}, 2021.

\bibitem[\protect\citeauthoryear{Fan \bgroup \em et al.\egroup
  }{2019}]{Fan2019StrategiesFS}
Angela Fan, Mike Lewis, and Yann Dauphin.
\newblock Strategies for structuring story generation.
\newblock In {\em ACL}, 2019.

\bibitem[\protect\citeauthoryear{Gibbs~Jr}{1994}]{gibbs1994poetics}
Raymond~W Gibbs~Jr.
\newblock {\em The poetics of mind: Figurative thought, language, and
  understanding}.
\newblock Cambridge University Press, 1994.

\bibitem[\protect\citeauthoryear{Goldfarb-Tarrant \bgroup \em et al.\egroup
  }{2020}]{goldfarb2020content}
Seraphina Goldfarb-Tarrant, Tuhin Chakrabarty, Ralph Weischedel, and Nanyun
  Peng.
\newblock Content planning for neural story generation with aristotelian
  rescoring.
\newblock In {\em the 2020 Conference on Empirical Methods in Natural Language
  Processing (EMNLP)}, 2020.

\bibitem[\protect\citeauthoryear{Han \bgroup \em et al.\egroup
  }{2019}]{han2019joint}
Rujun Han, Qiang Ning, and Nanyun Peng.
\newblock Joint event and temporal relation extraction with shared
  representations and structured prediction.
\newblock In {\em Proceedings of the 2019 Conference on Empirical Methods in
  Natural Language Processing and the 9th International Joint Conference on
  Natural Language Processing (EMNLP-IJCNLP)}, pages 434--444, 2019.

\bibitem[\protect\citeauthoryear{Han \bgroup \em et al.\egroup
  }{2020}]{han2020knowledge}
Rujun Han, Yichao Zhou, and Nanyun Peng.
\newblock Domain knowledge empowered structured neural net for end-to-end event
  temporal relation extraction.
\newblock In {\em the 2020 Conference on Empirical Methods in Natural Language
  Processing (EMNLP)}. Association for Computational Linguistics, 2020.

\bibitem[\protect\citeauthoryear{Han \bgroup \em et al.\egroup
  }{2021}]{han2021econet}
Rujun Han, Xiang Ren, and Nanyun Peng.
\newblock Econet: Effective continual pretraining of language models for event
  temporal reasoning.
\newblock In {\em The 2021 Conference on Empirical Methods in Natural Language
  Processing (EMNLP)}, 2021.

\bibitem[\protect\citeauthoryear{Han \bgroup \em et al.\egroup
  }{2022}]{han2022go}
Rujun Han, Hong Chen, Yufei Tian, and Nanyun Peng.
\newblock Go back in time: Generating flashbacks in stories with event plots
  and temporal prompts.
\newblock In {\em 2022 Annual Conference of the North American Chapter of the
  Association for Computational Linguistics (NAACL)}, 2022.

\bibitem[\protect\citeauthoryear{He \bgroup \em et al.\egroup
  }{2019}]{he2019pun}
He~He, Nanyun Peng, and Percy Liang.
\newblock Pun generation with surprise.
\newblock In {\em 2019 Annual Conference of the North American Chapter of the
  Association for Computational Linguistics (NAACL-HLT 2019)}, volume~1, 2019.

\bibitem[\protect\citeauthoryear{Hovy and Spruit}{2016}]{hovy2016social}
D.~Hovy and S.~L. Spruit.
\newblock The social impact of natural language processing.
\newblock In {\em Association for Computational Linguistics (ACL)}, 2016.

\bibitem[\protect\citeauthoryear{Hsu \bgroup \em et al.\egroup
  }{2022a}]{hsu2021discourse}
I-Hung Hsu, Xiao Guo, Premkumar Natarajan, and Nanyun Peng.
\newblock Discourse-level relation extraction via graph pooling.
\newblock In {\em The Thirty-Sixth AAAI Conference On Artificial Intelligence
  Workshop on Deep Learning on Graphs: Method and Applications (DLG-AAAI)},
  2022.

\bibitem[\protect\citeauthoryear{Hsu \bgroup \em et al.\egroup
  }{2022b}]{hsu2022degree}
I-Hung Hsu, Kuan-Hao Huang, Elizabeth Boschee, Scott Miller, Prem Natarajan,
  Kai-Wei Chang, and Nanyun Peng.
\newblock Degree: A data-efficient generation-based event extraction model.
\newblock In {\em 2022 Annual Conference of the North American Chapter of the
  Association for Computational Linguistics (NAACL)}, 2022.

\bibitem[\protect\citeauthoryear{Huang \bgroup \em et al.\egroup
  }{2019}]{huang2019learning}
Xiao Huang, Li~Dong, Elizabeth Boschee, and Nanyun Peng.
\newblock Learning a unified named entity tagger from multiple partially
  annotated corpora for efficient adaptation.
\newblock In {\em The 2019 SIGNLL Conference on Computational Natural Language
  Learning (CoNLL)}, 2019.

\bibitem[\protect\citeauthoryear{Huang \bgroup \em et al.\egroup
  }{2021}]{huang2021tempgen}
Kung-Hsiang Huang, Sam Tang, and Nanyun Peng.
\newblock Document-level entity-based extraction as template generation.
\newblock In {\em The 2021 Conference on Empirical Methods in Natural Language
  Processing (EMNLP)}, 2021.

\bibitem[\protect\citeauthoryear{Justo \bgroup \em et al.\egroup
  }{2014}]{justo2014extracting}
Raquel Justo, Thomas Corcoran, Stephanie~M Lukin, Marilyn Walker, and
  M~In{\'e}s Torres.
\newblock Extracting relevant knowledge for the detection of sarcasm and
  nastiness in the social web.
\newblock {\em Knowledge-Based Systems}, 69:124--133, 2014.

\bibitem[\protect\citeauthoryear{Lewis \bgroup \em et al.\egroup
  }{2020}]{lewis2020bart}
Mike Lewis, Yinhan Liu, Naman Goyal, Marjan Ghazvininejad, Abdelrahman Mohamed,
  Omer Levy, Veselin Stoyanov, and Luke Zettlemoyer.
\newblock {BART}: Denoising sequence-to-sequence pre-training for natural
  language generation, translation, and comprehension.
\newblock In {\em Proceedings of the 58th Annual Meeting of the Association for
  Computational Linguistics}, pages 7871--7880, 2020.

\bibitem[\protect\citeauthoryear{Lu and Peng}{2021}]{lu2021efficient}
Sidi Lu and Nanyun Peng.
\newblock On efficient training, controllability and compositional
  generalization of insertion-based language generators.
\newblock {\em arXiv preprint arXiv:2102.11008}, 2021.

\bibitem[\protect\citeauthoryear{Martin \bgroup \em et al.\egroup
  }{2017}]{martin2017event}
Lara Martin, Prithviraj Ammanabrolu, William Hancock, Shruti Singh, Brent
  Harrison, and Mark Riedl.
\newblock Event representations for automated story generation with deep neural
  nets.
\newblock In {\em Proceedings of the Thirty-Second AAAI Conference on
  Artificial Intelligence}, 2017.

\bibitem[\protect\citeauthoryear{Meehan}{1976}]{meehan1976metanovel}
James~Richard Meehan.
\newblock The metanovel: writing stories by computer.
\newblock Technical report, YALE UNIV NEW HAVEN CONN DEPT OF COMPUTER SCIENCE,
  1976.

\bibitem[\protect\citeauthoryear{Mittal \bgroup \em et al.\egroup
  }{2022}]{Mittal2022ambipun}
Anirudh Mittal, Yufei Tian, and Nanyun Peng.
\newblock Ambipun: Generating humorous puns with ambiguous context.
\newblock In {\em 2022 Annual Conference of the North American Chapter of the
  Association for Computational Linguistics (NAACL), short}, 2022.

\bibitem[\protect\citeauthoryear{Peng \bgroup \em et al.\egroup
  }{2017}]{Peng16graphlstm}
Nanyun Peng, Hoifung Poon, Chris Quirk, Kristina Toutanova, and Wen-Tau Yih.
\newblock Cross-sentence n-ary relation extraction with graph {LSTM}.
\newblock {\em Transaction of the Association for Computational Linguistics
  (TACL)}, 2017.

\bibitem[\protect\citeauthoryear{Radford \bgroup \em et al.\egroup
  }{2019}]{radford2019language}
Alec Radford, Jeffrey Wu, Rewon Child, David Luan, Dario Amodei, Ilya
  Sutskever, et~al.
\newblock Language models are unsupervised multitask learners.
\newblock {\em OpenAI blog}, 1(8):9, 2019.

\bibitem[\protect\citeauthoryear{Shah \bgroup \em et al.\egroup
  }{2020}]{shah2020predictive}
D.~S. Shah, H.~A. Schwartz, and D.~Hovy.
\newblock Predictive biases in natural language processing models: A conceptual
  framework and overview.
\newblock In {\em Association for Computational Linguistics (ACL)}, 2020.

\bibitem[\protect\citeauthoryear{Sheng \bgroup \em et al.\egroup
  }{2019}]{sheng2019woman}
Emily Sheng, Kai-Wei Chang, Premkumar Natarajan, and Nanyun Peng.
\newblock The woman worked as a babysitter: On biases in language generation.
\newblock In {\em 2019 Conference on Empirical Methods in Natural Language
  Processing (EMNLP), short}, 2019.

\bibitem[\protect\citeauthoryear{Sheng \bgroup \em et al.\egroup
  }{2020}]{sheng2020towards}
Emily Sheng, Kai-Wei Chang, Premkumar Natarajan, and Nanyun Peng.
\newblock Towards controllable biases in language generation.
\newblock In {\em the 2020 Conference on Empirical Methods in Natural Language
  Processing (EMNLP)-Findings, long}, 2020.

\bibitem[\protect\citeauthoryear{Sheng \bgroup \em et al.\egroup
  }{2021}]{sheng2021nice}
Emily Sheng, Kai-Wei Chang, Premkumar Natarajan, and Nanyun Peng.
\newblock "nice try, kiddo": Ad hominems in dialogue systems.
\newblock In {\em The 2021 Annual Conference of the North American Chapter of
  the Association for Computational Linguistics (NAACL)}. Association for
  Computational Linguistics, 2021.

\bibitem[\protect\citeauthoryear{Stein \bgroup \em et al.\egroup
  }{2018}]{stein2018using}
Elizabeth Stein, Robert~A Pearlman, and Tyler~P Tate.
\newblock Using metaphors and figurative language to improve communication
  about patient suffering.
\newblock {\em Journal of Pain and Symptom Management}, 56(6):e68--e69, 2018.

\bibitem[\protect\citeauthoryear{Stowe \bgroup \em et al.\egroup
  }{2021}]{stowe2021metaphor}
Kevin Stowe, Tuhin Chakrabarty, Nanyun Peng, Smaranda Muresan, and Iryna
  Gurevych.
\newblock Metaphor generation with conceptual mappings.
\newblock In {\em Proceedings of the Conference of the 59th Annual Meeting of
  the Association for Computational Linguistics (ACL)}, 2021.

\bibitem[\protect\citeauthoryear{Tian and Peng}{2022}]{tian2022sonnet}
Yufei Tian and Nanyun Peng.
\newblock Zero-shot sonnet generation with discourse-level coherence and poetic
  features.
\newblock In {\em 2022 Annual Conference of the North American Chapter of the
  Association for Computational Linguistics (NAACL)}, 2022.

\bibitem[\protect\citeauthoryear{Yao \bgroup \em et al.\egroup
  }{2019}]{yao2019plan}
Lili Yao, Nanyun Peng, Weischedel Ralph, Kevin Knight, Dongyan Zhao, and Rui
  Yan.
\newblock {Plan-And-Write}: Towards better automatic storytelling.
\newblock In {\em The Thirty-Third AAAI Conference on Artificial Intelligence
  (AAAI-19)}, 2019.

\bibitem[\protect\citeauthoryear{Yoshimura \bgroup \em et al.\egroup
  }{2015}]{yoshimura2015simile}
Eriko Yoshimura, Misako Imono, Seiji Tsuchiya, and Hirokazu Watabe.
\newblock A simile recognition system using a commonsense sensory association
  method.
\newblock {\em Procedia Computer Science}, 60:55--62, 2015.

\end{thebibliography}

\end{document}